\documentclass{article}
\usepackage{nlprs01}
\usepackage{epsf}
\title{Universal Model for Paraphrasing\\ --- Using Transformation Based on a Defined Criteria ---}
\author{Masaki Murata and Hitoshi Isahara\\
Communications Research Laboratory \\
{\normalsize 2-2-2 Hikaridai, Seika-cho, Soraku-gun, Kyoto, 619-0289, Japan}\\
{\normalsize \{murata,isahara\}@crl.go.jp}}

\begin{document}
\maketitle
\begin{abstract}
This paper describes a universal model for paraphrasing 
that transforms according to defined criteria. 
We showed that 
by using different criteria 
we could construct different kinds of paraphrasing systems 
including one for answering questions, 
one for compressing sentences, 
one for polishing up, 
and one for transforming 
written language to spoken language. 
\end{abstract}

\section{Introduction}

The term ``paraphrasing'' used in this paper means 
the process of rewriting sentences 
without altering the meaning. 
It includes 
generating easy sentences from difficult ones, 
polished sentences from broken or poor ones, or 
sentences used in spoken language from ones used in written language, 
which is useful for generating speech from written texts. 
Moreover, generating concise sentences that have almost the same meaning 
as their long, tedious original, 
which is classified under summarization, 
is also a type of paraphrasing. 
Paraphrasing is useful for many natural-language processing techniques,  
and it is important for generating these techniques. 

This paper describes a universal model that achieves 
many kinds of paraphrasing. 
It transforms sentences according to 
a predefined criteria. We show that our model can handle 
many kinds of paraphrases 
by selecting the most appropriate criteria 
for each type of paraphrasing. 
Since this model includes several of criteria 
for different types of paraphrasing, 
it can deal with a whole range of 
paraphrasing types. 

\section{Paraphrasing model}

\begin{figure}[t]
      \begin{center}
      \epsfile{file=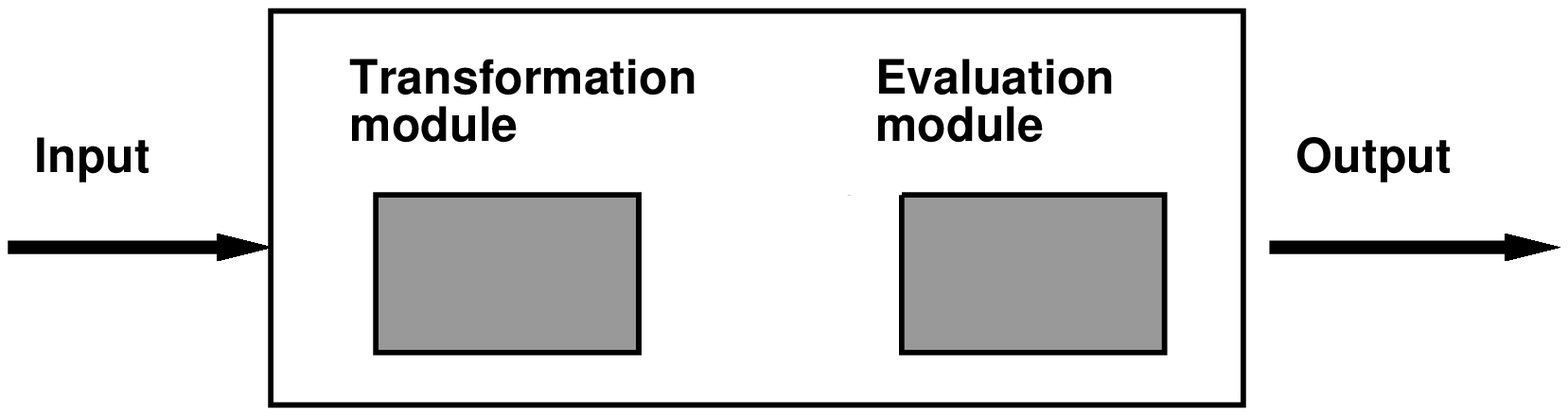,height=3cm,width=8cm} 
      \end{center}
      \caption{Our paraphrasing model}
      \label{fig:model}
\end{figure}

Figure \ref{fig:model} shows 
our paraphrasing model. 
It consists of 
two modules: a transformation and 
an evaluation module. 
A sentence that requires transformation 
is input into the system. 
Several potential transformation types are generated in 
the transformation module 
and then tested in the evaluation module, 
where most appropriate type is selected. 
It is then used for the transformation 
and the result is output. 

{
\begin{itemize}
\item 
  Transformation module
  
  This module generates the potential transformation types. 
  They are based on hand-written rules, 
  rules automatically detected by machines, 
  or on a combination of both. 

\item 
  Evaluation module

  This module selects the most appropriate transformation type 
  by using predefined criteria. 
  The criteria needs to 
  be adapted each time according to the particular 
  problem it should handle. 

\end{itemize}}

Here are several example criteria used 
in the evaluation module: 
{
\begin{itemize}
\item 
  Similarity 

  To establish the similarity between X and Y, 
  we first suppose that 
  all the rewriting rules in the transformation module 
  comply with the restriction 
  that the transformation does not alter the meaning. 
  We then transform X and Y in the transformation module 
  so that they are similar as possible. 
  Then we calculate their similarity correctly 
  even if X and Y expressed the same meanings differently. 

\item 
  Length

  To compress a sentence without 
  changing the meaning, 
  similarly to the sentence-compression process, 
  which is classified under summarization, 
  we again 
  suppose that 
  all the rewriting rules in the transformation module 
  comply with the restriction that 
  transformation does not alter the meaning. 
  Here, we use the length of the sentence as 
  the main criterion of transformation. 
  We can compress a sentence 
  by repeatedly transforming input sentences 
  to decrease their lengths.

\item 
  Frequency (or probability of occurrence)\footnote{Probabilities of occurrence
    in corpora have been used in many studies 
    on spelling-error correction and generation \cite{Spell_check,brown,Ratnaparkhi_nlg_naacl2000}.}

  To polish poor or unconnected sentences 
  also all the rewriting rules in the transformation module 
  have to comply with 
  the restriction that 
  transformation does not alter 
  the meaning.  
  We transform sentences that require 
  polishing according to 
  how frequently part of these sentences appear 
  in the corpora, so they 
  can be transformed into more sophisticated ones. 
  We can explain this by using an easier example: 
  if the input data include the word ``summarisation'' 
  and the transformation module has 
  a function that changes ``summarisation'' to ``summarization,'' 
  we count how often both ``summarisation'' and ``summarization'' 
  have so far appeared 
  by using Penn Treebank or another corpus. 
  When the frequency of ``summarization'' exceeds 
  that of ``summarisation,'' 
  we change ``summarisation'' in the input data 
  to ``summarization.'' 

  Furthermore, 
  we can use several types of corpora 
  for measuring the frequency or 
  calculating the probability, 
  and they have different results. 
  For example, when 
  input data in written language and 
  corpora comprising of spoken language is used, 
  the input data is converted into spoken language. 
  Thus, if we have input data that include ``is not'' 
  and the transformation module has 
  a function that changes ``is not'' to ``isn't'', 
  since ``isn't'' occurs more frequently than ``is not'' 
  in the spoken language corpus, 
  ``is not'' is changed to ``isn't.'' 

  Now, suppose that the input data are sentences 
  that have difficult expressions such as those used in legal documents. 
  When we use a set of easy sentences for the corpora 
  to measure their frequency, 
  the difficult sentences are transformed into easy ones. 
  Or, suppose that 
  the input data is a novel written by 
  an unknown writer and 
  a set of materials written by Shakespeare is used 
  for the corpora that measures the frequency. 
  In this case, a new novel in Shakespearean style 
  would be output.\footnote{Moreover, our model can handle machine translation \cite{brown}
    by applying translation rules in the transformation module 
    and using corpora written in the target language 
    for calculating the probabilities.}

\item 
  Judging the grammatical validity of a sentence 

  Measuring the frequency can be 
  used to polish sentences; 
  therefore, it can also 
  be used to judge whether a sentence is grammatical or not. 
  But when the criteria is too restricting 
  for establishing the grammatical validity, 
  we can instead use only one of the following: 

{
  \begin{itemize}
  \item 
    The expressions used in the transformed version should 
    occur at least once in the corpora. 
    (This measure is often used 
    in spell-check systems \cite{Spell_check}.)

  \item 
    The probability of occurrence in the corpora should exceed a certain threshold. 

  \item 
    The probability of occurrence
    in the corpora should be higher than 
    that when the surroundings are not used for calculating 
    the probabilities. 

  \end{itemize}
}

\begin{table*}[t]
  \begin{center}
    \caption{Sentences in the database (English translation of Japanese sentences)}
    \label{tab:mensetsu}
    \leavevmode
\begin{tabular}[h]{|p{15cm}|}\hline
\baselineskip=0.85\baselineskip 
Usually, when a Japanese person hears that an American lives in New York, 
he or she thinks that the American lives in New York City. 
This is a common mistake, however. 
New York City takes up only a very small area of southern New York State. 
It takes about eight hours to drive from New York City to Niagara Falls, 
which is also in New York State. 
The majority of the state consists of mountains, forests, fields, rivers, 
lakes, and swamps. The people who live in these central and northern areas 
of the state usually live in small towns. 
Farming is the most common occupation among these New York State 
residents, and corn is the most common crop grown by them. 
\\\hline
\end{tabular}
  \end{center}
\end{table*}
  
  The criteria we described here are 
  more similar to conditions, 
  and would be most effective when combined with 
  other criteria. 
  We should use these criteria additionally 
  when other criteria cannot guarantee 
  the grammatical validation of a sentence 
  in a transformation. 

\item 
  Judging the equivalence in meaning of a sentence 
  before and after transformation 

  When we do not know 
  whether the transformation in the transformation module 
  comply with the restriction that 
  the transformation should not alter 
  the meaning of a sentence, 
  this criteria is required. 
  However, we doubt that 
  equivalence in meaning can be 
  judged at all. 

  For ad hoc solutions 
  we can apply the following two methods, 
  either separately or simultaneously: 
{
  \begin{itemize}
  \item 
    We check the transformation rules by hand 
    and only use those that satisfy 
    the meaning-equivalence criteria. 
    And/or, we list the cases when 
    the rules satisfy the meaning-equivalence by hand 
    and the ones that do not, and 
    then judge the meaning-equivalence 
    by using that particular data. 

  \item 
    We extract only those rules that reliably 
    satisfy the meaning-equivalence 
    when they are extracted automatically. 
    And/or, we use a machine that learns 
    the conditions when 
    the rules satisfy the meaning-equivalence 
    are also learned automatically 
    when rules are extracted automatically. 

  \end{itemize}
}

This item is more similar to a condition than a criteria, 
as in the previous item. 
It is used in addition to other measures.\footnote{We can 
imagine further criteria for similar conditions, 
such as that 
the sentence should have at most seven phrases, plus or minus two, 
whose modifiees are not determined \cite{miller56,yngve60,Murata_7pm2_art}.}
    
\end{itemize}}
We can imagine other measures than the ones we described.\footnote{If research 
on polite expressions or easily understandable expressions 
can successfully produce criteria, it should enable 
automatic transformation to polite or 
easily understandable expressions. 
The same result can be achieved by using 
corpora that include only polite or 
only easily understandable sentences for measuring the probabilities of occurrences.}

In the following sections, 
we will demonstrate 
what kinds of criteria we applied for transformation 
in our model by using concrete examples from our research. 

\begin{table*}[t]
  \begin{center}
    \leavevmode
    \caption{Examples of the QA system}
    \label{tab:qa_result}
\begin{tabular}{|r|l|p{12.5cm}|}\hline
\multicolumn{1}{|c|}{Sim.} & \multicolumn{1}{|c|}{} & \multicolumn{1}{|c|}{Sentences}\\\hline
32.1 & quest. & The most general occupation among the residents of central and northern New York State is X. \\
32.1 & data &  Farming is the most common occupation among these New York State 
residents, and corn is the most common crop grown by them. \\\hline
103.1 & data & Farming is the most general occupation among these New York State 
residents, and corn is the most common crop grown by them. \\
82.5& data & Farming is the most common occupation among these residents of New York State, and corn is the most common crop grown by them.\\
... & ... & ...\\\hline
186.5 & data & Farming is the most general occupation among these residents of New York State, and corn is the most common crop grown by them.\\\
... & ... & ...\\\hline
... & ... & ...\\\hline
219.5 & quest. & X is the most general occupation among the residents of central and northern New York State. \\
219.5 & data & Farming is the most general occupation among these residents of New York State \\\hline
& Ans. & = Farming\\
& Sup. & = Farming is the most general occupation among these residents of New York State  \\\hline
\end{tabular}
\end{center}
\end{table*}

\begin{table}[t]
  \begin{center}
    \leavevmode
    \caption{transformation rule used in the transformation module}
    \label{tab:hitode_kisoku}
\begin{tabular}[h]{|lcl|}\hline
\baselineskip=0.7\baselineskip 
X is Y & $\Leftrightarrow$ & Y is X \\
general & $\Leftrightarrow$ & common\\
these X residents & $\Leftrightarrow$ & these residents of X\\
, X & $\Leftrightarrow$ & \\\hline
\end{tabular}
  \end{center}
\end{table}

\section{The case of a QA system}
\label{sec:qa}

Our question-answering system takes the following procedures.\footnote{There 
are many studies on the QA system \cite{MURAX,trec8qa}.}
\footnote{Our question-answering system uses original 
  paraphrasing techniques. 
  These techniques produce 
  very accurate question-answering results 
  because the system detects answers 
  when the transformed question sentence and the transformed data sentence 
  are very similar. }
\begin{enumerate}
\item 
  The system extracts 
  sentences, including the answer of the question sentence, 
  from sentences in the database. 

\item 
  The system rewrites 
  the extracted sentences and the question sentence 
  so that they are as similar as possible. 

\item 
  The system compares the rewritten sentences from the database 
  to the rewritten question sentence. 
  It then outputs the phrase in the rewritten sentence from the database 
  that corresponds to the interrogative pronoun 
  in the rewritten question sentence as the answer. 

\end{enumerate}

For example, 
when we are given the data \cite{eiken2k_eng} shown in Table \ref{tab:mensetsu}, 
we are asked 
the question of 
``What is the most general occupation among the residents of central and northern New York State?'' 
Our system\footnote{Our actual system is in Japanese and 
the example sentence in this paper is the English translation 
of the Japanese sentence.} 
transforms the question sentence into 
a declarative sentence and 
the interrogative pronoun is changed to X. 
The sentence most similar to this question sentence is extracted, 
and the system takes the state shown in the first line of Table \ref{tab:qa_result}. 
We suppose that 
our system applies the transformation rule shown in Table \ref{tab:hitode_kisoku}. 
The question sentence and the extracted sentences are rewritten 
to increase the similarity between them. 
Finally, as shown in the table, the similarity reaches a maximum level of 219.5 and 
cannot be increased. 
At this stage, the system compares 
the sentence from the database and the question sentence 
and detects ``Farming'' easily 
by extracting the phrases in the sentence from the database 
that correspond to X. 

\begin{figure*}[t]

  \begin{center}
\fbox{
  \begin{minipage}[h]{15cm}

    \fbox{Definition of ``reverse'' in Dictionary A:}

  \begin{tabular}[h]{l@{ }l@{ }l@{ }l@{ }l@{ }l@{ }l@{ }l@{ }l@{ }l}
    {\it junjo} & , & {\it ichi} & {\it nado} & {\it -no} & {\it kankei} & {\it -ga} & {\it sakasama-ni} & {\it irekawat} &  {\it -teiru}\\
    (order) & (,) & (location) & (etc.) & (of) & (relation) & {\sf nom} & (upside-down) & (change places) & (-ing)\\
    \multicolumn{10}{l}{(The relationship of the order, location and so on is changed upside-down.)}\\
  \end{tabular}\\[0.1cm]

  \fbox{Definition of ``reverse'' in Dictionary B:}

  \begin{tabular}[h]{llllllll}
    {\it junjo} & , & {\it ichi} & , & {\it kankei} &{\it -ga} & {\it hikkuri-kaet} &  {\it -teiru}\\
    (order) & (,) & (location) & (,) & (relation) & {\sf nom} & (be overturned) & (-ing)\\
    \multicolumn{8}{l}{(The relationship of the order and location is overturned.)}\\
  \end{tabular}\\[0.1cm]

  \fbox{Results of comparing the two definitions}

  \begin{tabular}[h]{lll|ll|ll|ll|l}
    {\it junjo} & , & {\it ichi} & {\it nado} & {\it -no} & {\it kankei}& {\it -ga} & {\it sakasama-ni} & {\it irekawat} &  {\it -teiru}\\
          &   &      & (etc.)& (of) &        &     & (upside-down) & (be changed) & \\
          &   &      & ,    &     &        &     & {\it hikkuri-kaet} & & \\
          &   &      & ,    &     &        &     & (be overturned) & & \\
  \end{tabular}

  \end{minipage}
}
  \end{center}

  \caption{Example of extracting rules for paraphrasing}
  \label{fig:ext_henkei}
\end{figure*}

Our QA system paraphrases 
by using similarity as a criterion. 
Because the system paraphrases sentences 
to increase their similarity, 
it facilitates comparing the question sentence and the data sentence. 
Here, we showed the QA system as an example. 
The similarity criterion can also be used for 
most cases of calculating similarity. 
For example, 
a high-level information retrieval system 
could determine the similarity of a query and 
a retrieval document after they are rewritten 
to ensure their similarity is the highest possible.\footnote{In an anaphora resolution \cite{murata_coling98}, 
the system cannot resolve the anaphora 
when the identity or inclusion-relationship 
of ``hole'' expressed by 
``a hole which is at the base of a huge cedar tree nearby'' 
and ``the hole at the base of the cedar tree'' 
cannot be judged. 
But when we rewrite them based on the similarity criterion and 
obtain 
``a hole at the base of a huge cedar tree nearby'' and ``the hole at the base of the cedar tree,'' 
the system understands that the former expression includes the latter one
(that the two trees, ``a huge cedar tree nearby'' and ``the cedar tree,'' are the same 
is determined easily because the former ``tree'' only has additional adjectives), 
and that the latter one refers to the former one.}

\section{The case of a sentence compression system}
\label{sec:compress}

Recently, many studies on summarization have been conducted. 
We also performed 
sentence compression \cite{Knight_sent_comp}, which is classified under 
summarization. 

We researched automatic extraction of rewriting rules 
from two different dictionaries. 
Here is a brief explanation of our research. 
Two Japanese dictionaries gave 
the definitions 
shown in Figure \ref{fig:ext_henkei} 
for the word {\it abekobe} meaning ``reverse''. 
We expected to 
extract the pairs of expressions having the same meaning 
by comparing the two definitions, 
since they both defined the same word 
and thus had the same meaning. 
We compared the two by using the unix command ``diff'' 
and obtain the results shown in the figure. 
From the results, we determined that 
{\it nado-no} ``etc.'' and ``,'' were interchangeable, as well as 
{\it sakasama-ni irekawatte} ``be changed upside-down'' 
and {\it hikkuri-kaet} ``be overturned''. 
We actually obtained 67,632 rewriting rules by using this method. 
However, they also included many incorrect ones. 
So we automatically selected rewriting rules 
that appear more than once in the comparison, 
because rewriting rules that appear twice or more in the comparison 
are more accurate.\footnote{In our actual experiments, 
we used probabilistic equations to detect rules. 
We cannot explain this method in detail in this paper. 
But the results would be similar to the results 
for detecting rules appearing more than once.}
The number of selected rules was 775. 
The research in this section uses these 775 rules 
as the rules in the transformation module. 

\begin{table*}[t]
  \begin{center}
    \leavevmode
    \caption{Example of sentence compression}
    \label{tab:compress_result}
\renewcommand{\arraystretch}{1}
\begin{tabular}{|l|}\hline
\multicolumn{1}{|c|}{Example of correctly transformed results}\\\hline
\begin{tabular}[h]{llllll}
{\it kokonoka} & \underline{\it -kara} & {\it -no} & {\it kankoku} & {\it houmon} & {\it -dewa}\\
(9th)    & (from)            & (of)& (Korea) & (visit) & (in)\\
\multicolumn{6}{l}{(in the visit to Korea of \underline{from} the 9th)}\\[0.1cm]
\end{tabular}\\
\begin{tabular}[h]{llllll}
{\it rekishi} & {\it -no} & \underline{\it nagare} & \underline{\it -no} & {\it naka} & {\it -de}\\
(history) & (of) & (flow) & (of) & (middle) & (in)\\
\multicolumn{6}{l}{(in the middle \underline{of the flow} of history)}\\[0.1cm]
\end{tabular}\\
\begin{tabular}[h]{llllll}
{\it juuoku} & {\it doru} & {\it no} & {\it tuika} & \underline{\it teki} & {\it sochi}\\
(a billion) & (dollar) & (of) & (supplement) & (-ary) & (step)\\
\multicolumn{6}{l}{(a supplement\underline{ary} step of one billion dollars)}\\
\end{tabular}\\\hline
\multicolumn{1}{|c|}{Example of incorrectly transformed results}\\\hline
\begin{tabular}[h]{lll}
{\it jiyuu} & \underline{\it to} & {\it minshushugi} \\
(liberty) & and & (democracy)\\
\multicolumn{3}{l}{(liberty and democracy)}\\[0.1cm]
\end{tabular}\\
\begin{tabular}[h]{lllllllll}
X-{\it san} & {\it wo} & {\it kouho} & {\it to-shite} & {\it youritsu} & \underline{\it suru} & \underline{\it koto} & {\it wo} & {\it kimeta} \\
(Mr. X) & {\sf obj} & (candidate) & (as) & (support) & (do) & {\sf obj} & (decide)\\
\multicolumn{8}{l}{(decided to support Mr. X as the candidate)}\\
\end{tabular}\\\hline
\end{tabular}
\end{center}
\end{table*}

Here we considered summarizing 
newspaper articles and 
used the following criteria in the evaluation module. 
{
  \begin{itemize}
\item 
  The transformed sentence should be shorter as possible. 

\item 
  The expressions in the transformed sentence should 
  appear at least once in the corpora, which contained 
  two-years' worth of newspaper articles, 
  to verify the grammatical validity of a sentence. 

\end{itemize}}

Strictly speaking, we used the following procedures. 
{
  \begin{enumerate}
\item 
  The system analyzed an input sentence morphologically 
  by using the Japanese morphological analyzer JUMAN \cite{JUMAN3.5_e} 
  and divided it into a string of morphemes.

\item 
  \label{enum:proc1}
  The system performed the following procedures 
  for each morpheme from left to right. 

  \begin{enumerate}
  \item 
    \label{enum:proc1__1}
    When the string of morphemes $S$, 
    whose first morpheme is the current one
    (including no morphemes, e.g., ``'') 
    matched 
    the $A_i$ string 
    from the transformation rule $R_i$ ($A_i$ $\Rightarrow$ $B_i$), 
    the $B_i$ string was extracted 
    as the candidate of the transformed expression. 
    We referred to the string of the $k$-gram morphemes 
    just before $S$ as $S1_i$ and 
    to the one just after as $S2_i$. 

  \item 
    \label{enum:proc1__2}
    The system counted the number 
    of the strings reduced 
    when string $A_i$ was changed to $B_i$ 
    against each $B_i$. 
    We referred to the $i$ 
    when the value was the highest as $m$. 
  
  \item 
    The system counted the frequency of 
    the string of $S1_m$$B_m$$S2_m$ 
    in the corpus used in the evaluation module. 
    When it occurred at least once, 
    the system transformed $A_m$ to $B_m$ 
    and performed the procedure on the next morphology. 
  \end{enumerate}
\end{enumerate}}
\noindent
where, $k$ is a constant. 

Here, we used a simple method using 
k-gram as the environments for calculating 
to facilitate the experiments. 
Since we used a simple method, 
we set $k$ at 2 to 
increase the precision rate and 
decrease the recall rate. 

We carried out the experiments on sentence compression 
using newspaper articles. 
An example of the results are shown in Table \ref{tab:compress_result}. 
The underlined part is 
the part that was removed in the transformation. 
Because this section focused on
sentence compression, 
transformation rules to remove strings 
were frequently used. 
The ``from,'' ``of flow,'' and {\it teki}
were appropriately deleted, 
which succeeded in compressing the sentences. 
But the results also included faulty deletions of 
{\it to} ``and,'' and {\it surukoto} (do). 
Omitting {\it to} changes the original meaning of 
{\it jiyuu to minshushugi}, ``liberty and democracy''. 
Omitting {\it surukoto} (do) changes 
a verb {\it youritsu} (support) into a noun, 
and the phrase X-{\it san wo} ``Mr. X'' is missing a verb. 
The sentence is therefore not grammatical. 
To correct these errors 
requires using a new evaluation method that includes 
syntactic features. 

We emphasize that we can compress sentences 
by using the sentence length as a criterion in the evaluation module, 
as our experiments confirmed. 

\section{The case of a sentence-polishing-up system}

In this section we describe 
a sentence-polishing-up system. 
We used the same 775 transformation rules 
as in the previous section. 

Here, we tried to polish sentences from newspaper articles and 
applied the following criteria in the evaluation module. 
{
\begin{itemize}
\item 
  The substrings of the transformed sentence 
  should occur frequently in the corpora, 
  which contained 
  two-years' worth of the news-paper articles.\footnote{We need to use 
    a corpus/corpora with the same subject as the target sentence 
    to polish it up accordingly.}
  \footnote{The research in this section is similar to 
    that on the spelling or word correction \cite{Spell_check}. 
    But in the cases of spelling or word correction, 
    ungrammatical sentences are input. 
    In contrast, in sentence-polishing-up systems, 
    grammatical sentences are input and 
    the systems make them more sophisticated. 
    Performing such sentence-polishing-up without 
    the rewriting rules 
    as we automatically extract is difficult.}

\end{itemize}}

We used the procedures 
by which we changed part of 2c 
in the procedures of Section \ref{sec:compress} 
to the following: 

\begin{enumerate}
\item[(c)] 
    The system counts the frequencies of 
    the strings of $S1_m$$A_m$$S2_m$ and $S1_m$$B_m$$S2_m$ 
    in the corpus used in the evaluation module. 
    When the number of the frequency of $S1_m$$B_m$$S2_m$ 
    exceeds that of $S1_m$$A_m$$S2_m$, 
    the system transforms $A_m$ to $B_m$ 
    and performs the procedure of the next morphology. 
  \end{enumerate}
  In this case, too, we used k=2.\footnote{\label{fn:kairyou} To 
    improve the results of transformation, 
    the frequency of each string $x$ in our procedures 
    must be changed to the probability of occurrence of $x$ 
  in the corpora when the given input data is used as the context. 
  Although our procedures use 
  the fixed $k*2$ morphemes of ``in front'' and ``behind'' as the context, 
  we should calculate the probabilities 
  by using the variable-length context and 
  more global information, such as syntactic information and 
  tense information, in the powerful probability-estimator 
  such as the maximum entropy method.}

\begin{table}[t]
\footnotesize
  \begin{center}
    \leavevmode
    \caption{Examples of results of sentence-polishing-up}
    \label{tab:suikou_result}
\begin{tabular}{|@{}l@{}|}\hline
\multicolumn{1}{|c|}{Example of correctly transformed results}\\\hline
\begin{tabular}[h]{lllll}
{\it kazoku} & , & {\it yuujin-ra} & {\it to} & {\it sugosu}\\
(family) & (,) & (friends) & (with) & (live)\\
         & {\it ya} &  & & \\
         & and &  & & \\
\multicolumn{5}{l}{(lives with her family, (, $\Rightarrow$ and) her friends.)}\\[0.4cm]
\end{tabular}\\
\begin{tabular}[h]{lll}
{\it shijiritsu} &    &{\it kaifuku} \\
(support rates) &  & (recovery) \\[-0.2cm]
 & {\it -no} &  \\
 & (of) &  \\[-0.1cm]
\multicolumn{3}{@{ }l@{}}{(support rates recovery ($\Rightarrow$ recovery of supports rates))}\\
\end{tabular}\\\hline
\multicolumn{1}{|c|}{Example of wrongly transformed results}\\\hline
\begin{tabular}[h]{llll}
8 & {\it pointo} &                 & {\it uwamawari}\\
8 & (points) &               & exceed\\[-0.2cm]
   &          &{\it  -mo} &\\
   &          & (no less than)&\\
\multicolumn{4}{l}{(exceed 8 points)}\\[0.4cm]
\end{tabular}\\
\begin{tabular}[h]{llll}
{\it sakunen} & {\it -wo} & {\it shoutyou} & {\it shita} \\
(last year) & {\sf obj} & (symbolize) & (did)\\
 &  &  & {\it suru} \\
 &  &  & (do)\\
\multicolumn{4}{l}{(symbolized last year)}\\
\end{tabular}\\\hline
\end{tabular}
\end{center}
\end{table}

We carried out the sentence-polishing experiments 
using newspaper articles. 
Some results are shown in Table \ref{tab:suikou_result}. 
The lower strings are the transformed ones. 
The results include sentences that were made 
more comprehensible 
by adding {\it ya} ``and'' and {\it no} ``of'', 
and others where 
{\it mo}, meaning ``no more than'' was added incorrectly. 
The latter transformation changed the meaning of the sentence. 
In another case the tense was changed incorrectly form 
{\it shita} ``did'' to {\it suru} ``do''. 
These errors resulted from errors in the automatic selection 
of the transformation rules. 

We want to stress that we can perform multiple kinds of paraphrasing 
by using various types of measures for an evaluation module. 
The results of this section included 
cases where the system enhanced the sentence lengths 
and modified the input sentences to 
be more comprehensible; 
the results were different from the ones obtained in the compression tests. 
These results thus confirm our assertation.

\begin{table*}[t]
\footnotesize
  \begin{center}
    \leavevmode
    \caption{Examples of transformation from 
      written to spoken language}
    \label{tab:s2p_henkei2}
\begin{tabular}{|l|}\hline
\begin{tabular}[h]{lllllllll}
sono & teigi & -wo & riyou & suru & toiu-koto & -ga &    & kangaerareru\\
(its) & (definition) & {\sf obj} & (use) & (do) & (that) & {\sf obj} &   & (think)\\
     &       &     &       &      &          &     & ma &             \\
     &       &     &       &      &          &     & {\sf filler} &             \\
\multicolumn{9}{l}{(We can think ({\it ma}) that we use its definition.)}\\[0.3cm]
   \end{tabular}\\
\begin{tabular}[h]{lllllllll}
dougi & hyougen & -wo & tyushutsu & suru &     & koto & -wo & kokoromiru.\\
(same-meaning) & (expression) & {\sf obj} & (extract) & (do) &   & (that) & {\sf obj} & (try)\\
 &  & &  & & toiu & & & \\
 &  & &  & & (that is/such)  & & & \\
\multicolumn{9}{l}{(We try (such) that we extract the same-meaning expressions.)}\\[0.3cm]
\end{tabular}\\
\begin{tabular}[h]{lllllllllll}
hindo & -de & souto & -shita & kekka &       &   & -wo & hyou & -ni & shimesu.\\
(frequency) & (by) & (sort) & (did) & (result) & & & {\sf obj} & (table) & (in) & (show)\\
      &     &       &        &       & toiu & -no  &     &      &     &        \\
      &     &       &        &       & (that is)&  (those)  &     &      &     &        \\
\multicolumn{11}{l}{(We show (those that is) the results sorted by frequency in the table.)}\\
\end{tabular}\\\hline
\end{tabular}
\end{center}
\end{table*}

\section{The case of a written-language-to-spoken-language transformation system}

Here, we tried to transform 
the sentences in written-language style 
to those in spoken-language style. 

We carried out new experiments 
for extracting the rules that transform 
strings from written to spoken language. 
These experiments were performed by using 
the same method as in Section \ref{sec:compress}
comparing the parallel corpora of written and spoken language. 
Our institution has been compiling these corpora. 
The written-language sentences are taken from academic papers 
and the spoken-language ones from their oral presentations. 
We obtained 72,835 rewriting rules from these experiments, 
but many contain the same incorrect rewriting rules as in 
Section \ref{sec:compress}. 
We thus automatically selected 240 rewriting rules 
that appear more than once. 
This section uses these rules in the transformation module.\footnote{We tried 
using the 775 rules in Section \ref{sec:compress} in addition to 
these 240 rules for the experiments of this section. 
However, the results were worse than if we had not used them. 
This is because the 775 rules included faulty transformation rules such as 
{\it suru} $\Rightarrow$ {\it shita} 
``do $\Rightarrow$ did''. 
We believe that if the 775 rules had not included such wrong rules, 
we could have used them as well for the experiments in this section.}

We used the following criteria in the evaluation module. 
{
  \begin{itemize}
\item 
  The substrings of the transformed sentence 
  should occur frequently in the spoke-language corpora.\footnote{These corpora 
    include 330,679 Japanese characters.}
  \footnote{We have to use 
    the same-domain corpora with the domain of the target to which input 
    data are transformed.}

\end{itemize}}

We followed the same procedure as in the previous section. 

We only changed from using a newspaper corpora 
to a spoken-language corpora. 
In the previous section, 
because the subject of the input data was same as 
that of the corpora used in the evaluation module, 
the system made the newspaper articles 
more similar in style to newspaper articles by polishing up. 
In contrast, in this section, 
because the input data was in written language and 
the corpus used in the evaluation module was spoken language, 
the data was made transformed 
from written to spoken language.\footnote{The research 
of this section is very similar to statistical machine translation \cite{brown}. 
In this section, the source language is written language and 
the target language is spoken language.}

We input sentences from one of our papers as experiments. 
The results are shown in Table \ref{tab:s2p_henkei2}. 
In these experiments, 
no incorrect transformation occurred. 
{\it Ma} is a filler and 
roughly means ``so-so.'' 
It is often used in spoken Japanese. 
{\it Toiu} means ``that is'' and is also used often 
in spoken Japanese. 
The transformed expressions in the table 
adequately produced 
the nuance of spoken Japanese. 
However, only a few transformed expressions were obtained 
and the transformation recall rate was low. 
Therefore, we need to improve the system (Footnote \ref{fn:kairyou}). 

We want to reiterate that we can perform multiple kinds of paraphrasing 
by applying various types of criteria in the evaluation module. 
In these experiments, we obtained expressions used in spoken language. 
So the results are sufficient to confirm our claim. 

\section{Conclusion}

We demonstrated that 
a method that transforms sentences based on certain criteria 
can be used 
as a universal model for paraphrasing. 
We showed four ways our systems can apply 
this simple model and 
confirmed that 
by using different criteria 
we could construct different systems, 
including question answering, 
sentence compression, 
sentence polishing-up, and 
written-language to spoken-language transformation. 

Implementing various types of paraphrases using 
a universal model has the following advantages:
\begin{itemize}
\item 
  Since some components in the model 
  can be used in different types of systems, 
  we can use them to construct more systems 
  after having constructed one. 

\item 
  We can construct a new paraphrasing system 
  by merely changing a small part of an existing system 
  (e.g. the criteria in the evaluation module). 
  Therefore, 
  we can construct new paraphrasing systems very easily.

\end{itemize}
In the future, 
we hope to construct 
more types of 
paraphrasing systems by using our universal model. 

{\small
\bibliographystyle{acl}
\bibliography{mysubmit}
}

\end{document}